\documentclass[runningheads]{llncs}
\usepackage[utf8]{inputenc}
\usepackage[T1]{fontenc}

\usepackage{import}
\usepackage{amsmath,graphicx}
\usepackage{times}
\usepackage[dvipsnames,table]{xcolor}
\usepackage{pgf}

\usepackage{tabularx,booktabs}
\usepackage{arydshln}
\usepackage{tabu}
\usepackage{adjustbox}
\usepackage{enumerate,multicol}
\usepackage{multirow}
\usepackage{multicol}
\usepackage{colortbl}
\usepackage{caption}
\usepackage{subcaption}
\usepackage{gensymb}
\usepackage{float}
\usepackage{balance}
\usepackage{xspace}
\usepackage{amssymb}
\usepackage{bbding,enumitem}
\usepackage{booktabs}
\usepackage{hyperref}
\usepackage{color, colortbl}
\hypersetup{colorlinks,linkcolor={red!50!black},citecolor={blue!50!black},urlcolor={black}}
\usepackage[table]{xcolor}
\usepackage{times}

\usepackage{soul}
\usepackage{url}

\usepackage{amsmath}
\usepackage{nicefrac, xfrac}
\usepackage{booktabs}

\definecolor{gray1}{RGB}{210,210,210}
\definecolor{gray2}{RGB}{50,50,50}
\definecolor{gray3}{RGB}{150,150,150}
\definecolor{RYB1}{RGB}{218,232,252}
\definecolor{RYB4}{RGB}{108,142,191}
\definecolor{blue2}{RGB}{230,240,250}

\makeatletter
\newcommand\avsuminner[2]{%
  {\sbox0{$\m@th#1\sum$}%
   \vphantom{\usebox0}%
   \ooalign{%
     \hidewidth
     \smash{\vrule height\dimexpr\ht0+1pt\relax depth\dimexpr\dp0+1pt\relax}%
     \hidewidth\cr
     $\m@th#1\sum$\cr
   }%
  }%
}
\definecolor{shadecolor}{rgb}{0.9,0.9,0.9}

\usepackage{verbatim}

\usepackage{tikz}

\newcommand{\DP}[2]{%
  \begin{tikzpicture}
    \fill[color=#2]   (0.0 , 0.0) rectangle (#1*7.8ex , 2ex );
  \end{tikzpicture}%
}
\newcommand{\DPD}[2]{%
  \begin{tikzpicture}
    \fill[color=#2]   (0.0 , 0.0) rectangle (#1*6.2ex , 2ex );
  \end{tikzpicture}%
}

\usepackage{calc}
\usepackage{stackengine}
\newcommand\clapp[3][0pt]{\stackengine{0pt}{#3}{\kern#1#2}{O}{c}{F}{F}{L}}

\usepackage{pgfplots}
\usepackage{pgfplotstable}
\usetikzlibrary{pgfplots.groupplots}
\usepgfplotslibrary{colorbrewer}
\pgfplotsset{/pgfplots/error bars/error bar style={black,thick}}

\pgfplotsset{compat=1.11,
        /pgfplots/ybar legend/.style={
        /pgfplots/legend image code/.code={%
        \draw[##1,/tikz/.cd,bar width=3pt,yshift=-0.2em,bar shift=0pt]
                plot coordinates {(0cm,0.8em)};},
},}

\usepackage{color}

\newcommand{\etal}[0]{et al. }
\begin{document}
\title{Event Recognition in Laparoscopic Gynecology Videos with Hybrid Transformers}
\titlerunning{Event Recognition in Laparoscopic Gynecology Videos}
%

\author{Sahar Nasirihaghighi\inst{1} \and 
Negin Ghamsarian\inst{2} \and
Heinrich Husslein\inst{3} \and
Klaus Schoeffmann\inst{1}}

\authorrunning{Nasirihaghighi, S., et al.}
%

\institute{Institute of Information Technology (ITEC), Klagenfurt University, Austria \and
Center for AI in Medicine, University of Bern, Switzerland \and
Department of Gynecology and Gynecological Oncology, Medical University Vienna, Austria}

\maketitle              
\begin{abstract}

Analyzing laparoscopic surgery videos presents a complex and multifaceted challenge, with applications including surgical training, intra-operative surgical complication prediction, and post-operative surgical assessment. Identifying crucial events within these videos is a significant prerequisite in a majority of these applications. In this paper, we introduce a comprehensive dataset tailored for relevant event recognition in laparoscopic gynecology videos. Our dataset includes annotations for critical events associated with major intra-operative challenges and post-operative complications. To validate the precision of our annotations, we assess event recognition performance using several CNN-RNN architectures. Furthermore, we introduce and evaluate a hybrid transformer architecture coupled with a customized training-inference framework to recognize four specific events in laparoscopic surgery videos. Leveraging the Transformer networks, our proposed architecture harnesses inter-frame dependencies to counteract the adverse effects of relevant content occlusion, motion blur, and surgical scene variation, thus significantly enhancing event recognition accuracy. Moreover, we present a frame sampling strategy designed to manage variations in surgical scenes and the surgeons' skill level, resulting in event recognition with high temporal resolution. We empirically demonstrate the superiority of our proposed methodology in event recognition compared to conventional CNN-RNN architectures through a series of extensive experiments.


\end{abstract}

\section{Introduction}
\label{sec:intro}

Laparoscopic surgery, an integral subset of minimally invasive procedures, has significantly transformed surgical practices by offering numerous benefits, such as reduced patient trauma, accelerated recovery time, and diminished postoperative discomfort~\cite{aldahoul2021transfer}. Laparoscopic surgeries entail performing procedures through small incisions using a camera-equipped laparoscope~\cite{lux2010novel,loukas2015smoke}. The laparoscope is inserted through one of the trocars, and the camera provides a high-definition view of the abdominal structures on a monitor. Laparoscopic videos capture a wealth of information critical for surgical evaluation, education, and research~\cite{loukas2018keyframe,loukas2018video}. Analyzing these videos encompasses diverse objectives, ranging from tool presence detection, surgical instrument tracking, and organ segmentation to surgical phase recognition and, crucially, surgical action or event recognition.

Surgical event recognition in laparoscopic gynecology surgery videos is a critical area of research within the field of computer-assisted intervention (CAI). Analyzing and recognizing surgical events in these videos holds great potential for improving surgical outcomes, enhancing training, and facilitating post-operative assessment~\cite{schoeffmann2015keyframe}. However, even in the presence of advanced CAI systems~\cite{czempiel2020tecno}, recognizing surgical events from laparoscopic surgery video recordings remains a formidable challenge. This complexity arises from the inherent variability in patient anatomy, the type of surgery, and surgeon skill levels~\cite{funke2019video}, coupled with hindering factors such as the smoke presence within the surgical area~\cite{leibetseder2017image}, blurred lenses due to fog or blood, objects occlusions, and camera motion~\cite{wang2018smoke}. In this matter, we present and publicly release a comprehensive dataset for deep-learning-based event recognition in laparoscopic gynecology videos. Our dataset features four critical events, including Abdominal Access, Bleeding, Coagulation/Transaction, and Needle Passing. Since event recognition plays a prominent role in laparoscopic workflow recognition, the utilization of this dataset can effectively contribute to accelerating surgical training and improving surgical outcomes.

The combinations of CNNs-RNNs have proven to be a powerful technique in analyzing surgical videos, as they effectively leverage joint spatio-temporal information \cite{ghamsarian2020enabling,ghamsarian2020relevance,ghamsarian2021lensid,ghamsarian2021relevance}. 
However, the evolution of event recognition has taken another significant turn with the recent adoption of transformer models~\cite{vaswani2017attention,dosovitskiy2020image}. Originally designed for machine translation, the transformers have undergone a paradigm shift in their application to various time-series modeling challenges~\cite{shi2022recognition}, including action recognition in laparoscopic surgery videos. The effectiveness of transformers stems from their remarkable self-attention mechanism, which enables capturing complex temporal relationships and dependencies within data sequences~\cite{bello2019attention}.
In this paper, we present a robust framework based on hybrid transformer models for relevant event recognition in laparoscopic gynecology videos to further enhance workflow analysis in this surgery.

This paper delivers the following key contributions:
\begin{itemize}
    \item We introduce and evaluate a comprehensive dataset tailored for event recognition in gynecology laparoscopic videos. The video dataset, along with our customized annotations, will be publicly released upon the acceptance of this paper.
    
    \url{https://ftp.itec.aau.at/datasets/LapGyn6-Events/} 
    \item We propose a hybrid transformer-based model designed to identify significant events within laparoscopic gynecology surgery.
    \item We comprehensively evaluate the performance of the proposed model against several combinations of CNN-RNN frameworks.
    
\end{itemize}

The subsequent sections of this paper are structured as follows: Section \ref{sec:relatedwork} provides a comprehensive review of the relevant literature, highlighting distinctions from the present work. Section \ref{sec:dataset} presents a detailed explanation of the dataset, offering a thorough insight into its composition and characteristics. Section \ref{sec:approach} delineates the proposed hybrid transformer-based framework for robust event recognition in laparoscopic gynecology surgery. The experimental setups are explained in Section \ref{sec:setups}, and the experimental results are presented in Section \ref{sec:results}. Finally, Section \ref{sec:conclusion} provides the conclusions drawn from the study.

\section{Related Work}
\label{sec:relatedwork}

This section provides a comprehensive overview of the latest advancements in laparoscopic surgical video analysis and action recognition within laparoscopy videos. Given the present study's focus on CNN-RNNs and Transformers, we aim to highlight methods that effectively utilize recurrent layers or Transformers for the analysis of laparoscopy videos.

\paragraph{Laparoscopic Surgery Video Analysis with CNN-RNNs:}
Tiwinanda \etal~\cite{twinanda2016endonet} introduce an architecture named EndoNet, crafted to concurrently perform phase recognition and tool presence detection tasks within a multi-task framework.
In~\cite{namazi2018automatic}, Namazi \etal introduced a surgical phase detection technique using deep learning system (SPD-DLS) to recognize surgical phases within laparoscopic videos. The approach employs a combination of the Convolutional Neural Network and Long Short-Term Memory model, which collectively analyzes both spatial and temporal information to detect surgical phases across video frames.
Jin \etal proposed a CNN-LSTM for joint surgical phase recognition and tool presence detection~\cite{jin2020multi}. While the CNN is responsible for extracting tool presence and spatial evidence of the surgical phase, the recurrent head is employed to model temporal dependencies between frame-wise feature maps. 
The study by Tobias et al.~\cite{czempiel2020tecno} presents a multi-stage temporal convolutional network (MS-TCN) tailored for surgical phase recognition. Termed TeCNO, derived from "Temporal Convolutional Networks for the Operating Room," this approach revolves around a surgical workflow recognition structure employing a two-stage TCN model, refining the extracted spatial features by establishing a comprehensive understanding of the current frame through an analysis of its preceding frames.
In LapTool-Net by Namazi \etal~\cite{namazi2022contextual}, Gated Recurrent Unit (GRU) is employed to harness inter-frame correlations for multi-label instrument presence classification in laparoscopic videos.
Golany \etal~\cite{golany2022artificial} suggest a multi-stage temporal convolution network for surgical phase recognition. In our previous work \cite{Nasirihaghighi2023}, we investigate action recognition in videos from laparoscopic gynecology with CNN-RNN model. In this paper, the detection of six actions—namely, abdominal access, anatomical grasping, knot pushing, needle pulling, needle pushing, and suction—from the gynecologic laparoscopy videos are explored by incorporating CNNs and cascaded bidirectional recurrent layers.

\paragraph{Action Recognition in Laparoscopy Videos with transformers:}

Huang \etal ~\cite{huang2022surgical} introduce a surgical action recognition model termed SA, leveraging the transformer's attention mechanism. This encoder-decoder framework simultaneously classifies current and subsequent actions. The encoder network employs self-attention and cross-attention to grasp the intra-frame and inter-frame contextual relationships, respectively. Subsequently, the decoder predicts future and current actions, utilizing cross-attention to relate surgical actions. In ~\cite{kiyasseh2023vision},  Kiyasseh \etal introduce a comprehensive surgical AI system (SAIS) capable of decoding various elements of intraoperative surgical activity from surgical videos. SAIS effectively segments surgical videos into distinct intraoperative activities.  
SAIS operates with two parallel streams processing separate input data types: RGB surgical videos and optical flow. Features from each frame are extracted using a pre-trained Vision Transformer (ViT) on ImageNet. These frame features then undergo transformation via a series of transformer encoders, yielding modality-specific video features. 
In another study, Shi \etal ~\cite{shi2022attention} present a novel method for surgical phase recognition, employing an attention-based spatial-temporal neural network. The approach leverages the attention mechanism within the spatial model, enabling adaptive integration of local features and global dependencies for enhanced spatial feature extraction. For temporal modeling, IndyLSTM and a non-local block are introduced to capture sequential frame features, establishing temporal dependencies among input clip frames.
Sharghi \etal~\cite{he2022empirical} introduce a dual-stage model comprising backbone and temporal components to address surgical activity recognition. The study highlights the effectiveness of the Swin-Transformer and BiGRU temporal model as a potent combination in achieving this goal.

\section{Laparascopic Gynecology Dataset}
\label{sec:dataset}

We possess a dataset comprising 174 laparoscopic surgery videos extracted from a larger pool of over 600 gynecologic laparoscopic surgeries recorded with a resolution of $1920\times 1080$ at the Medical University of Vienna. All videos are annotated by clinical experts into Abdominal Access, Grasping Anatomy, Transection, Coagulation, Knotting, Knot Pushing, Needle Passing, String Cutting, Suction, Irrigation, as well as the adverse event of Bleeding. 
While numerous actions or events are executed during laparoscopic surgery, we have specifically extracted four pivotal events of particular interest to physicians: (I) Abdominal Access, (II) Bleeding, (III) Coagulation and Transection, and (IV) Needle Passing. We detail these relevant events in the following.

\begin{description}
    \item[Abdominal Access.] The initial and one of the most critical steps in laparoscopic surgery involves achieving abdominal access by directly inserting a trocar with a blade into the abdominal cavity. This step holds particular significance, as it is the point where a substantial portion of potentially life-threatening complications, amounting to approximately $(40- 50)\%$ of cases, can arise. These complications encompass critical issues like damage to major blood vessels, urinary tract injuries, and bowel damage \cite{AAiLS}. While typically occurring as the first event, Abdominal Access can also take place at various points during the surgery. 

    \item[Bleeding.] Intraoperative hemorrhage stands as an adverse event potentially arising due to various factors, such as inadvertent cutting during the surgical process. Although some instances of bleeding during laparoscopic surgery are inevitable, they remain relatively rare and are proficiently addressed by the adept surgical team. Although laparoscopic surgery is specifically designed to minimize bleeding through the utilization of small incisions and precise techniques, bleeding events are probable. These events are paramount in laparoscopic surgery due to potentially affecting patient safety, surgical precision, and overall surgical outcomes. 

    \item[Coagulation and Transection.] These two events are frequently intertwined in laparoscopic surgery. Coagulation serves a crucial role in controlling bleeding at the transection site. This dual action ensures not only a clear surgical field but also a well-controlled surgical environment, allowing for precise removal of pathological tissues. 

    \item[Needle Passing.] This event encompasses the delicate movement of a needle into and out of the tissue, often conducted in tasks like suturing and tissue manipulation. Needle passing in laparoscopic surgery demands a particular skill set and a high level of precision, as it involves maneuvering the needle through tissue structures. Suturing, in particular, emerges as a cornerstone of laparoscopic surgery. Closing incisions with precise care can control bleeding to maintain a clear surgical field, repair damaged tissues and organs, and ultimately affect the overall success of the procedure. Beyond its technical significance, suturing in laparoscopy is a surgical skill assessment step~\cite{lim2017laparoscopic}.

    \end{description}

\begin{table}[bt!]
\centering
\caption{Visualization of relevant event annotations for ten representative laparoscopic videos from our dataset.}
\label{tbl:statistics_events_2}
\resizebox{1\textwidth}{!}{%
\begin{tabular}{m{1.0cm}m{12cm}}
\specialrule{.12em}{.05em}{.05em}
Case & Events \\\midrule
1&\DP{0.15}{Dandelion}\DP{2.08}{Gray}\DP{0.01}{Gray}\DP{-0.01}{Gray}\DP{0.01}{ForestGreen}\DP{0.10}{Gray}\DP{0.02}{Gray}\DP{-0.02}{Gray}\DP{0.02}{ForestGreen}\DP{0.09}{Gray}\DP{0.04}{Gray}\DP{-0.04}{Gray}\DP{0.04}{ForestGreen}\DP{0.04}{Gray}\DP{0.00}{Gray}\DP{-0.00}{Gray}\DP{0.00}{ForestGreen}\DP{0.09}{Gray}\DP{0.01}{Gray}\DP{-0.01}{Gray}\DP{0.01}{ForestGreen}\DP{0.86}{Gray}\DP{0.01}{Gray}\DP{-0.01}{Gray}\DP{0.01}{ForestGreen}\DP{0.03}{Gray}\DP{0.01}{Gray}\DP{-0.01}{Gray}\DP{0.01}{ForestGreen}\DP{0.27}{Gray}\DP{0.01}{Gray}\DP{-0.00}{Gray}\DP{0.00}{ForestGreen}\DP{0.37}{Gray}\DP{0.00}{Gray}\DP{-0.00}{Gray}\DP{0.00}{ForestGreen}\DP{0.07}{Gray}\DP{0.01}{Gray}\DP{-0.01}{Gray}\DP{0.01}{ForestGreen}\DP{0.18}{Gray}\DP{0.00}{Gray}\DP{-0.00}{Gray}\DP{0.00}{ForestGreen}\DP{0.11}{Gray}\DP{0.00}{Gray}\DP{-0.00}{Gray}\DP{0.00}{ForestGreen}\DP{1.15}{Gray}\DP{0.02}{Gray}\DP{-0.02}{Gray}\DP{0.02}{ForestGreen}\DP{0.05}{Gray}\DP{0.01}{Gray}\DP{-0.01}{Gray}\DP{0.01}{ForestGreen}\DP{2.76}{Gray}\DP{0.01}{Gray}\DP{-0.01}{Gray}\DP{0.01}{ForestGreen}\DP{0.02}{Gray}\DP{0.01}{Gray}\DP{-0.01}{Gray}\DP{0.01}{ForestGreen}\DP{0.06}{Gray}\DP{0.00}{Gray}\DP{-0.00}{Gray}\DP{0.00}{ForestGreen}\DP{1.03}{Gray}\DP{0.01}{Gray}\DP{-0.01}{Gray}\DP{0.01}{ForestGreen}\DP{0.07}{Gray}\DP{0.01}{Gray}\DP{-0.01}{Gray}\DP{0.01}{ForestGreen}\DP{0.08}{Gray}\DP{0.01}{Gray}\DP{-0.01}{Gray}\DP{0.01}{ForestGreen}\DP{0.11}{Gray}\DP{0.03}{Gray}\DP{-0.03}{Gray}\DP{0.03}{ForestGreen}\\
2&\DP{0.01}{Gray}\DP{0.04}{Gray}\DP{0.00}{Gray}\DP{0.12}{Gray}\DP{0.01}{Gray}\DP{-0.01}{Gray}\DP{0.01}{ForestGreen}\DP{0.17}{Gray}\DP{0.01}{Gray}\DP{-0.01}{Gray}\DP{0.01}{ForestGreen}\DP{0.26}{Gray}\DP{0.01}{Gray}\DP{-0.01}{Gray}\DP{0.01}{ForestGreen}\DP{0.08}{Gray}\DP{0.02}{Gray}\DP{-0.02}{Gray}\DP{0.02}{ForestGreen}\DP{0.14}{Gray}\DP{0.02}{Gray}\DP{-0.02}{Gray}\DP{0.02}{ForestGreen}\DP{0.14}{Gray}\DP{0.02}{Gray}\DP{-0.02}{Gray}\DP{0.02}{ForestGreen}\DP{0.56}{Gray}\DP{0.00}{Gray}\DP{-0.00}{Gray}\DP{0.00}{ForestGreen}\DP{0.14}{Gray}\DP{0.01}{Gray}\DP{-0.01}{Gray}\DP{0.01}{ForestGreen}\DP{0.17}{Gray}\DP{0.01}{Gray}\DP{-0.01}{Gray}\DP{0.01}{ForestGreen}\DP{0.10}{Gray}\DP{0.01}{Gray}\DP{-0.01}{Gray}\DP{0.01}{ForestGreen}\DP{0.03}{Gray}\DP{0.01}{Gray}\DP{-0.01}{Gray}\DP{0.01}{ForestGreen}\DP{0.02}{Gray}\DP{0.00}{Gray}\DP{-0.00}{Gray}\DP{0.00}{ForestGreen}\DP{0.02}{Gray}\DP{0.00}{Gray}\DP{-0.00}{Gray}\DP{0.00}{ForestGreen}\DP{0.03}{Gray}\DP{0.01}{Gray}\DP{-0.01}{Gray}\DP{0.01}{ForestGreen}\DP{0.02}{Gray}\DP{0.00}{Gray}\DP{-0.00}{Gray}\DP{0.00}{ForestGreen}\DP{0.02}{Gray}\DP{0.00}{Gray}\DP{-0.00}{Gray}\DP{0.00}{ForestGreen}\DP{0.14}{Gray}\DP{0.04}{Maroon}\DP{0.06}{Gray}\DP{0.01}{Gray}\DP{-0.01}{Gray}\DP{0.01}{ForestGreen}\DP{0.03}{Gray}\DP{0.01}{Gray}\DP{-0.01}{Gray}\DP{0.01}{ForestGreen}\DP{0.12}{Gray}\DP{0.01}{Gray}\DP{-0.01}{Gray}\DP{0.01}{ForestGreen}\DP{0.16}{Gray}\DP{0.01}{Gray}\DP{-0.01}{Gray}\DP{0.01}{ForestGreen}\DP{0.02}{Gray}\DP{0.01}{Gray}\DP{-0.01}{Gray}\DP{0.01}{ForestGreen}\DP{1.14}{Gray}\DP{0.05}{ForestGreen}\DP{0.07}{Gray}\DP{0.05}{ForestGreen}\DP{0.04}{Gray}\DP{0.06}{ForestGreen}\DP{0.04}{Gray}\DP{0.05}{ForestGreen}\DP{0.09}{Gray}\DP{0.06}{ForestGreen}\DP{0.11}{Gray}\DP{0.04}{ForestGreen}\DP{0.41}{Gray}\DP{0.05}{ForestGreen}\DP{0.07}{Gray}\DP{0.04}{ForestGreen}\DP{0.06}{Gray}\DP{0.05}{ForestGreen}\DP{0.06}{Gray}\DP{0.07}{ForestGreen}\DP{0.14}{Gray}\DP{0.05}{ForestGreen}\DP{0.03}{Gray}\DP{0.04}{ForestGreen}\DP{0.03}{Gray}\DP{0.04}{ForestGreen}\DP{0.03}{Gray}\DP{0.06}{ForestGreen}\DP{0.04}{Gray}\DP{0.07}{Maroon}\DP{-0.06}{Gray}\DP{0.05}{ForestGreen}\DP{0.03}{Gray}\DP{0.04}{ForestGreen}\DP{0.08}{Gray}\DP{0.04}{ForestGreen}\DP{0.05}{Gray}\DP{0.05}{Maroon}\DP{0.18}{Gray}\DP{0.05}{ForestGreen}\DP{1.04}{Gray}\DP{0.11}{ForestGreen}\DP{-0.01}{Gray}\DP{0.03}{Maroon}\DP{0.03}{Gray}\DP{0.09}{ForestGreen}\DP{0.03}{Gray}\DP{0.04}{ForestGreen}\DP{0.17}{Gray}\DP{0.23}{ForestGreen}\DP{0.07}{Gray}\DP{0.07}{ForestGreen}\DP{0.10}{Gray}\DP{0.09}{ForestGreen}\DP{0.06}{Gray}\DP{0.06}{ForestGreen}\DP{0.05}{Gray}\DP{0.07}{ForestGreen}\DP{0.04}{Gray}\DP{0.01}{ForestGreen}\DP{0.92}{Gray}\DP{0.08}{ForestGreen}\DP{-0.04}{Gray}\DP{0.01}{Maroon}\DP{0.05}{Gray}\DP{0.08}{ForestGreen}\DP{-0.06}{Gray}\DP{0.02}{Maroon}\DP{0.06}{Gray}\DP{0.03}{ForestGreen}\DP{0.01}{Gray}\DP{0.00}{ForestGreen}\\
3&\DP{0.05}{ForestGreen}\DP{0.82}{Gray}\DP{0.06}{ForestGreen}\DP{0.15}{Gray}\DP{0.05}{ForestGreen}\DP{0.12}{Gray}\DP{0.04}{ForestGreen}\DP{0.19}{Gray}\DP{0.06}{ForestGreen}\DP{0.46}{Gray}\DP{0.01}{Gray}\DP{-0.01}{Gray}\DP{0.01}{ForestGreen}\DP{0.16}{Gray}\DP{0.01}{Gray}\DP{-0.01}{Gray}\DP{0.01}{ForestGreen}\DP{0.17}{Gray}\DP{0.03}{Gray}\DP{-0.03}{Gray}\DP{0.03}{ForestGreen}\DP{0.10}{Gray}\DP{0.03}{Gray}\DP{-0.03}{Gray}\DP{0.03}{ForestGreen}\DP{0.14}{Gray}\DP{0.03}{Gray}\DP{-0.03}{Gray}\DP{0.03}{ForestGreen}\DP{0.33}{Gray}\DP{0.01}{Gray}\DP{-0.01}{Gray}\DP{0.01}{ForestGreen}\DP{0.02}{Gray}\DP{0.00}{Gray}\DP{-0.00}{Gray}\DP{0.00}{ForestGreen}\DP{0.02}{Gray}\DP{0.00}{Gray}\DP{-0.00}{Gray}\DP{0.00}{ForestGreen}\DP{0.01}{Gray}\DP{0.01}{Gray}\DP{-0.00}{Gray}\DP{0.00}{ForestGreen}\DP{0.01}{Gray}\DP{0.00}{Gray}\DP{-0.00}{Gray}\DP{0.00}{ForestGreen}\DP{0.01}{Gray}\DP{0.00}{Gray}\DP{-0.00}{Gray}\DP{0.00}{ForestGreen}\DP{0.23}{Gray}\DP{0.00}{Gray}\DP{-0.00}{Gray}\DP{0.00}{ForestGreen}\DP{0.01}{Gray}\DP{0.00}{Gray}\DP{-0.00}{Gray}\DP{0.00}{ForestGreen}\DP{0.02}{Gray}\DP{0.00}{Gray}\DP{-0.00}{Gray}\DP{0.00}{ForestGreen}\DP{0.07}{Gray}\DP{0.01}{Gray}\DP{-0.01}{Gray}\DP{0.01}{ForestGreen}\DP{0.01}{Gray}\DP{0.01}{Gray}\DP{-0.01}{Gray}\DP{0.01}{ForestGreen}\DP{0.07}{Gray}\DP{0.01}{Gray}\DP{-0.01}{Gray}\DP{0.01}{ForestGreen}\DP{-0.00}{Gray}\DP{0.08}{Maroon}\DP{0.04}{Gray}\DP{0.00}{Gray}\DP{-0.00}{Gray}\DP{0.00}{ForestGreen}\DP{0.01}{Gray}\DP{0.00}{Gray}\DP{-0.00}{Gray}\DP{0.00}{ForestGreen}\DP{0.21}{Gray}\DP{0.02}{Gray}\DP{-0.02}{Gray}\DP{0.02}{ForestGreen}\DP{0.11}{Gray}\DP{0.00}{Gray}\DP{-0.00}{Gray}\DP{0.00}{ForestGreen}\DP{1.25}{Gray}\DP{0.07}{ForestGreen}\DP{0.04}{Gray}\DP{0.06}{ForestGreen}\DP{0.05}{Gray}\DP{0.04}{ForestGreen}\DP{2.30}{Gray}\DP{0.04}{ForestGreen}\DP{0.07}{Gray}\DP{0.05}{ForestGreen}\DP{0.20}{Gray}\DP{0.05}{ForestGreen}\DP{0.22}{Gray}\DP{0.04}{ForestGreen}\DP{0.17}{Gray}\DP{0.04}{ForestGreen}\DP{0.10}{Gray}\DP{0.04}{ForestGreen}\DP{0.08}{Gray}\DP{0.06}{ForestGreen}\DP{0.17}{Gray}\DP{0.05}{ForestGreen}\DP{0.02}{Gray}\DP{0.31}{Maroon}\DP{0.42}{Gray}\DP{0.05}{ForestGreen}\\
4&\DP{0.05}{Dandelion}\DP{-0.02}{Gray}\DP{0.04}{Dandelion}\DP{1.39}{Gray}\DP{0.03}{Dandelion}\DP{0.71}{Gray}\DP{0.02}{Dandelion}\DP{2.53}{Gray}\DP{0.15}{ForestGreen}\DP{0.27}{Gray}\DP{0.12}{ForestGreen}\DP{0.31}{Gray}\DP{0.11}{ForestGreen}\DP{0.28}{Gray}\DP{0.12}{ForestGreen}\DP{0.46}{Gray}\DP{0.08}{ForestGreen}\DP{0.11}{Gray}\DP{0.02}{Gray}\DP{0.10}{Gray}\DP{0.10}{ForestGreen}\DP{0.77}{Gray}\DP{0.14}{ForestGreen}\DP{0.34}{Gray}\DP{0.14}{ForestGreen}\DP{0.38}{Gray}\DP{0.01}{Gray}\DP{0.92}{Gray}\DP{0.01}{Gray}\DP{-0.01}{Gray}\DP{0.01}{ForestGreen}\DP{0.04}{Gray}\DP{0.01}{Gray}\DP{-0.01}{Gray}\DP{0.01}{ForestGreen}\DP{0.03}{Gray}\DP{0.00}{Gray}\DP{-0.00}{Gray}\DP{0.00}{ForestGreen}\DP{0.16}{Gray}\DP{0.01}{Gray}\DP{-0.01}{Gray}\DP{0.01}{ForestGreen}\DP{0.04}{Gray}\DP{0.01}{Gray}\DP{-0.01}{Gray}\DP{0.01}{ForestGreen}\\
5&\DP{0.00}{Dandelion}\DP{0.94}{Gray}\DP{0.00}{Dandelion}\DP{0.47}{Gray}\DP{0.01}{Gray}\DP{-0.01}{Gray}\DP{0.01}{ForestGreen}\DP{0.04}{Gray}\DP{0.01}{Gray}\DP{-0.01}{Gray}\DP{0.01}{ForestGreen}\DP{0.09}{Gray}\DP{0.00}{Gray}\DP{-0.00}{Gray}\DP{0.00}{ForestGreen}\DP{0.03}{Gray}\DP{0.00}{Gray}\DP{-0.00}{Gray}\DP{0.00}{ForestGreen}\DP{0.02}{Gray}\DP{0.01}{Gray}\DP{-0.01}{Gray}\DP{0.01}{ForestGreen}\DP{0.03}{Gray}\DP{0.00}{Gray}\DP{-0.00}{Gray}\DP{0.00}{ForestGreen}\DP{0.38}{Gray}\DP{0.02}{ForestGreen}\DP{0.04}{Gray}\DP{0.01}{ForestGreen}\DP{0.04}{Gray}\DP{0.02}{ForestGreen}\DP{0.07}{Gray}\DP{0.02}{ForestGreen}\DP{0.07}{Gray}\DP{0.02}{ForestGreen}\DP{0.06}{Gray}\DP{0.02}{ForestGreen}\DP{0.05}{Gray}\DP{0.02}{ForestGreen}\DP{0.13}{Gray}\DP{0.01}{ForestGreen}\DP{0.06}{Gray}\DP{0.02}{ForestGreen}\DP{0.07}{Gray}\DP{0.01}{ForestGreen}\DP{0.03}{Gray}\DP{0.03}{ForestGreen}\DP{0.02}{Gray}\DP{0.01}{ForestGreen}\DP{0.03}{Gray}\DP{0.00}{ForestGreen}\DP{0.03}{Gray}\DP{0.02}{ForestGreen}\DP{0.01}{Gray}\DP{0.01}{ForestGreen}\DP{0.06}{Gray}\DP{0.01}{ForestGreen}\DP{0.05}{Gray}\DP{0.01}{ForestGreen}\DP{0.01}{Gray}\DP{0.01}{ForestGreen}\DP{0.08}{Gray}\DP{0.01}{ForestGreen}\DP{0.24}{Gray}\DP{0.02}{ForestGreen}\DP{0.01}{Gray}\DP{0.01}{ForestGreen}\DP{0.03}{Gray}\DP{0.02}{ForestGreen}\DP{0.07}{Gray}\DP{0.02}{ForestGreen}\DP{0.01}{Gray}\DP{0.02}{ForestGreen}\DP{0.17}{Gray}\DP{0.02}{ForestGreen}\DP{0.01}{Gray}\DP{0.01}{ForestGreen}\DP{0.05}{Gray}\DP{0.02}{ForestGreen}\DP{0.02}{Gray}\DP{0.01}{ForestGreen}\DP{0.14}{Gray}\DP{0.01}{ForestGreen}\DP{0.07}{Gray}\DP{0.01}{ForestGreen}\DP{0.03}{Gray}\DP{0.02}{ForestGreen}\DP{0.04}{Gray}\DP{0.01}{ForestGreen}\DP{0.12}{Gray}\DP{0.00}{ForestGreen}\DP{0.10}{Gray}\DP{0.00}{ForestGreen}\DP{0.05}{Gray}\DP{0.02}{ForestGreen}\DP{0.21}{Gray}\DP{0.02}{ForestGreen}\DP{0.02}{Gray}\DP{0.02}{ForestGreen}\DP{0.02}{Gray}\DP{0.01}{ForestGreen}\DP{0.01}{Gray}\DP{0.02}{ForestGreen}\DP{0.01}{Gray}\DP{0.01}{ForestGreen}\DP{0.06}{Gray}\DP{0.02}{Maroon}\DP{0.07}{Gray}\DP{0.02}{ForestGreen}\DP{0.01}{Gray}\DP{0.01}{ForestGreen}\DP{0.01}{Gray}\DP{0.02}{ForestGreen}\DP{0.03}{Gray}\DP{0.02}{ForestGreen}\DP{0.01}{Gray}\DP{0.04}{Maroon}\DP{0.02}{Gray}\DP{0.01}{Maroon}\DP{0.40}{Gray}\DP{0.02}{ForestGreen}\DP{0.00}{Gray}\DP{0.19}{Maroon}\DP{0.56}{Gray}\DP{0.10}{ForestGreen}\DP{0.00}{Gray}\DP{0.02}{ForestGreen}\DP{0.09}{Gray}\DP{0.06}{ForestGreen}\DP{0.02}{Gray}\DP{0.04}{ForestGreen}\DP{-0.01}{Gray}\DP{0.05}{Maroon}\DP{-0.00}{Gray}\DP{0.03}{ForestGreen}\DP{0.07}{Gray}\DP{0.01}{ForestGreen}\DP{0.03}{Gray}\DP{0.01}{ForestGreen}\DP{0.02}{Gray}\DP{0.01}{ForestGreen}\DP{0.01}{Gray}\DP{0.00}{ForestGreen}\DP{0.01}{Gray}\DP{0.01}{ForestGreen}\DP{0.12}{Gray}\DP{0.02}{ForestGreen}\DP{0.04}{Gray}\DP{0.08}{ForestGreen}\DP{0.03}{Gray}\DP{0.06}{ForestGreen}\DP{0.31}{Gray}\DP{0.05}{ForestGreen}\DP{0.03}{Gray}\DP{0.04}{ForestGreen}\DP{0.01}{Gray}\DP{0.08}{ForestGreen}\DP{0.06}{Gray}\DP{0.03}{ForestGreen}\DP{1.05}{Gray}\DP{0.02}{ForestGreen}\DP{0.02}{Gray}\DP{0.02}{ForestGreen}\DP{0.01}{Gray}\DP{0.02}{ForestGreen}\DP{0.02}{Gray}\DP{0.01}{ForestGreen}\DP{0.17}{Gray}\DP{0.01}{ForestGreen}\DP{0.01}{Gray}\DP{0.02}{ForestGreen}\DP{0.01}{Gray}\DP{0.01}{ForestGreen}\DP{0.38}{Gray}\DP{0.00}{ForestGreen}\DP{0.01}{Gray}\DP{0.00}{ForestGreen}\DP{0.01}{Gray}\DP{0.00}{ForestGreen}\DP{0.21}{Gray}\DP{0.00}{ForestGreen}\DP{0.00}{Gray}\DP{0.00}{ForestGreen}\\6&\DP{0.00}{Dandelion}\DP{0.24}{Gray}\DP{0.01}{Dandelion}\DP{0.34}{Gray}\DP{0.00}{Dandelion}\DP{0.34}{Gray}\DP{0.01}{Dandelion}\DP{0.23}{Gray}\DP{0.00}{Dandelion}\DP{0.16}{Gray}\DP{0.00}{Gray}\DP{3.84}{Gray}\DP{0.03}{Maroon}\DP{4.77}{Gray}\DP{0.01}{Dandelion}\\
7&\DP{0.18}{RoyalBlue}\DP{0.22}{Gray}\DP{0.11}{RoyalBlue}\DP{0.18}{Gray}\DP{0.01}{Gray}\DP{0.36}{Gray}\DP{0.19}{RoyalBlue}\DP{0.14}{Gray}\DP{0.09}{RoyalBlue}\DP{0.25}{Gray}\DP{0.13}{RoyalBlue}\DP{0.09}{Gray}\DP{0.07}{RoyalBlue}\DP{0.22}{Gray}\DP{0.95}{RoyalBlue}\DP{-0.29}{Gray}\DP{0.11}{Maroon}\DP{0.20}{Gray}\DP{0.28}{RoyalBlue}\DP{0.09}{Gray}\DP{0.01}{Maroon}\DP{0.04}{Gray}\DP{0.01}{Maroon}\DP{0.01}{Gray}\DP{0.00}{Maroon}\DP{0.06}{Gray}\DP{0.08}{Maroon}\DP{-0.05}{Gray}\DP{0.20}{RoyalBlue}\DP{-0.02}{Gray}\DP{0.00}{Maroon}\DP{0.41}{Gray}\DP{0.52}{RoyalBlue}\DP{0.18}{Gray}\DP{0.07}{RoyalBlue}\DP{0.27}{Gray}\DP{0.20}{RoyalBlue}\DP{0.25}{Gray}\DP{0.24}{RoyalBlue}\DP{-0.01}{Gray}\DP{0.06}{Maroon}\DP{0.74}{Gray}\DP{0.20}{RoyalBlue}\DP{0.27}{Gray}\DP{0.17}{RoyalBlue}\DP{0.27}{Gray}\DP{0.07}{RoyalBlue}\DP{0.12}{Gray}\DP{0.01}{Gray}\DP{2.01}{Gray}\DP{0.02}{Dandelion}\\
8&\DP{0.10}{Dandelion}\DP{1.10}{Gray}\DP{0.28}{RoyalBlue}\DP{0.30}{Gray}\DP{0.22}{RoyalBlue}\DP{0.28}{Gray}\DP{0.03}{Gray}\DP{0.70}{Gray}\DP{0.64}{RoyalBlue}\DP{0.35}{Gray}\DP{0.51}{RoyalBlue}\DP{0.38}{Gray}\DP{0.13}{RoyalBlue}\DP{0.35}{Gray}\DP{0.45}{RoyalBlue}\DP{0.42}{Gray}\DP{0.49}{RoyalBlue}\DP{0.32}{Gray}\DP{0.17}{RoyalBlue}\DP{0.27}{Gray}\DP{0.14}{RoyalBlue}\DP{0.74}{Gray}\DP{0.14}{RoyalBlue}\DP{0.39}{Gray}\DP{0.01}{Gray}\DP{0.92}{Gray}\DP{0.07}{Gray}\DP{0.03}{Gray}\DP{0.09}{Gray}\\
9&\DP{0.04}{Dandelion}\DP{-0.01}{Gray}\DP{0.05}{Dandelion}\DP{2.21}{Gray}\DP{0.46}{RoyalBlue}\DP{0.34}{Gray}\DP{0.02}{Gray}\DP{0.72}{Gray}\DP{0.08}{RoyalBlue}\DP{0.54}{Gray}\DP{0.25}{RoyalBlue}\DP{0.39}{Gray}\DP{0.30}{RoyalBlue}\DP{0.59}{Gray}\DP{0.22}{RoyalBlue}\DP{0.11}{Gray}\DP{0.08}{RoyalBlue}\DP{0.69}{Gray}\DP{0.46}{RoyalBlue}\DP{0.51}{Gray}\DP{0.39}{RoyalBlue}\DP{0.63}{Gray}\DP{0.01}{Gray}\DP{0.82}{Gray}\DP{0.02}{Gray}\DP{0.01}{Gray}\DP{0.08}{Gray}\\
10&\DP{0.01}{Dandelion}\DP{0.03}{Gray}\DP{0.00}{Dandelion}\DP{1.91}{Gray}\DP{0.01}{Dandelion}\DP{1.17}{Gray}\DP{0.00}{Gray}\DP{0.12}{Gray}\DP{0.02}{RoyalBlue}\DP{0.01}{Gray}\DP{0.01}{RoyalBlue}\DP{0.02}{Gray}\DP{0.03}{RoyalBlue}\DP{0.10}{Gray}\DP{0.03}{Gray}\DP{0.02}{Gray}\DP{0.00}{Gray}\DP{1.28}{Gray}\DP{0.01}{Gray}\DP{0.74}{Gray}\DP{0.03}{RoyalBlue}\DP{0.01}{Gray}\DP{0.02}{RoyalBlue}\DP{0.02}{Gray}\DP{0.03}{RoyalBlue}\DP{-0.01}{Gray}\DP{0.01}{Maroon}\DP{0.03}{Gray}\DP{0.02}{RoyalBlue}\DP{0.09}{Gray}\DP{0.03}{Gray}\DP{0.02}{Gray}\DP{0.00}{Gray}\DP{0.12}{Gray}\DP{0.04}{Gray}\DP{0.01}{Gray}\DP{0.01}{Gray}\DP{0.31}{Gray}\DP{0.02}{RoyalBlue}\DP{0.02}{Gray}\DP{0.02}{RoyalBlue}\DP{0.12}{Gray}\DP{0.04}{Gray}\DP{0.01}{Gray}\DP{0.00}{Gray}\DP{0.11}{Gray}\DP{0.02}{RoyalBlue}\DP{0.07}{Gray}\DP{0.06}{RoyalBlue}\DP{0.02}{Gray}\DP{0.01}{RoyalBlue}\DP{0.23}{Gray}\DP{0.00}{Gray}\DP{0.09}{Gray}\DP{0.02}{RoyalBlue}\DP{0.02}{Gray}\DP{0.02}{RoyalBlue}\DP{0.16}{Gray}\DP{0.00}{Gray}\DP{0.05}{Gray}\DP{0.00}{Gray}\DP{0.03}{Gray}\DP{0.00}{Gray}\DP{0.87}{Gray}\DP{0.07}{RoyalBlue}\DP{0.01}{Gray}\DP{0.03}{RoyalBlue}\DP{0.02}{Gray}\DP{0.01}{RoyalBlue}\DP{0.01}{Gray}\DP{0.02}{RoyalBlue}\DP{0.11}{Gray}\DP{0.01}{Gray}\DP{0.01}{Gray}\DP{0.00}{Gray}\DP{0.08}{Gray}\DP{0.01}{RoyalBlue}\DP{0.03}{Gray}\DP{0.02}{RoyalBlue}\DP{0.02}{Gray}\DP{0.02}{RoyalBlue}\DP{0.01}{Gray}\DP{0.04}{RoyalBlue}\DP{0.02}{Gray}\DP{0.04}{RoyalBlue}\DP{0.05}{Gray}\DP{0.02}{RoyalBlue}\DP{0.10}{Gray}\DP{0.01}{Gray}\DP{0.02}{Gray}\DP{0.00}{Gray}\DP{0.56}{Gray}\DP{0.01}{Gray}\DP{0.02}{Gray}\DP{0.00}{Gray}\DP{0.10}{Gray}\DP{0.00}{Gray}\DP{0.07}{Gray}\DP{0.00}{Gray}\DP{0.01}{Gray}\DP{0.00}{Gray}\DP{0.01}{Gray}\DP{0.00}{Gray}\DP{0.07}{Gray}\DP{0.00}{Gray}\\
\midrule
\end{tabular}
}
\resizebox{1\textwidth}{!}{%
\begin{tabular}{m{1.5cm}m{13 cm}}
Colormap &  Abd. Access \DPD{0.9}{Dandelion}, 
Bleeding \DPD{0.9}{Maroon},
Coag./Tran. \DPD{0.9}{ForestGreen},
Needle Passing \DPD{0.9}{RoyalBlue},
Irrelevant \DPD{0.9}{Gray}
\\
\specialrule{.12em}{.05em}{.05em}
\end{tabular}
}
\end{table}

\begin{figure*}[t!]
    \centering
    \begin{subfigure}[t]{0.24\textwidth}
    \centering
        \includegraphics[width=0.95\textwidth]{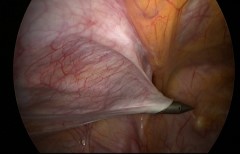}
    \end{subfigure}
    \begin{subfigure}[t]{0.24\textwidth}
    \centering
        \includegraphics[width=0.95\textwidth]{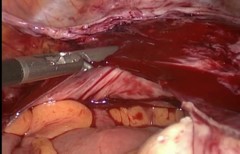}
    \end{subfigure}
    \begin{subfigure}[t]{0.24\textwidth}
    \centering
        \includegraphics[width=0.95\textwidth]{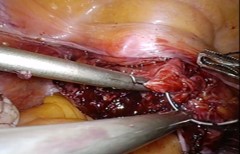}
    \end{subfigure}
    \begin{subfigure}[t]{0.24\textwidth}
    \centering
        \includegraphics[width=0.95\textwidth]{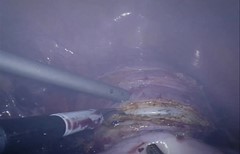}
    \end{subfigure}
    \\
    \begin{subfigure}[t]{0.24\textwidth}
    \centering
        \includegraphics[width=0.95\textwidth]{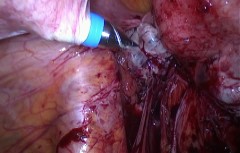}
        \caption*{Abdominal Access}    
    \end{subfigure}
    \begin{subfigure}[t]{0.24\textwidth}
    \centering
        \includegraphics[width=0.95\textwidth]{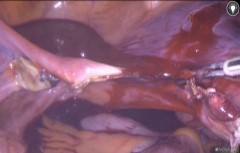}
        \caption*{Bleeding}    
    \end{subfigure}
    \begin{subfigure}[t]{0.24\textwidth}
    \centering
        \includegraphics[width=0.95\textwidth]{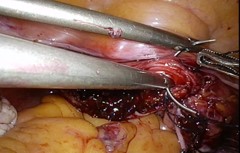}
        \caption*{Needle Passing}    
    \end{subfigure}
    \begin{subfigure}[t]{0.24\textwidth}
    \centering
        \includegraphics[width=0.95\textwidth]{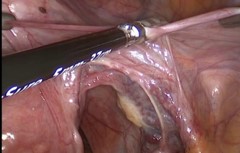}
        \caption*{Coag./Tran.}    
    \end{subfigure}
    \caption{Exemplar frames extracted from laparoscopic surgery videos.}
    \label{fig:phases-visualization}
\end{figure*}

\begin{table}[t!]
\centering
\caption{Statistics corresponding to the four different relevant annotated cases in our dataset.}
\label{tbl:duration_1}
\resizebox{\textwidth}{!}{
\begin{tabular}{l*{5}{>{\centering\arraybackslash}m{2.3cm}}}
    \specialrule{.12em}{.05em}{.05em}  
          & Event & No. Cases & No. Segments &  Segment Duration (sec) & Total duration (sec)\\
    \specialrule{.12em}{.05em}{.05em}  
         E1 & Abd. Access & 111 & 178 & 1 - 11 & 329.84 \\
    
         \rowcolor{shadecolor}E2 & Bleeding & 41 & 81 & 2 - 108 & 1577.20 \\

         E3 & Coag./Tran. & 12 & 584 & 2 - 43 & 2929.54 \\
         
         \rowcolor{shadecolor}E4 & Needle passing & 48 & 510 & 2 - 110 & 7036.43 \\

    \specialrule{.12em}{.05em}{.05em}
    
\end{tabular}
}
\end{table}
 Table~\ref{tbl:statistics_events_2} visualizes annotations for representative videos from our laparoscopic surgery dataset. Each row in the table corresponds to a distinct case of a laparoscopic surgery video. These cases exhibit varying durations, spanning from 11 minutes to approximately three hours.  Within every individual case, the presence of distinct events, including Abdominal Access, Bleeding, Coagulation/Transection, and Needle Passing, along with irrelevant events in this study, is depicted using different colors. We should emphasize that not all laparoscopic videos contain all relevant events. The segments exhibit varying durations, ranging from three seconds to over one minute in some cases. The details regarding the cases in each event are presented in Table ~\ref{tbl:duration_1}. It's worth noting that the segments with durations of less than one second are excluded from the training set as they do not meet the criteria for our proposed \textit{input dropout} strategy. Fig. ~\ref{fig:phases-visualization} depicts representative frames corresponding to these relevant events in laparoscopic gynecology surgery.

\section{Proposed Approach}
\label{sec:approach}

Within this section, we provide a comprehensive explanation of our proposed hybrid transformer model for the recognition of events in laparoscopic surgery videos. Additionally, we delineate our specialized training and inference framework.

\subsection{Hybrid transformer model}
\begin{figure*}[t!]
    \centering
    \includegraphics[width=12cm]{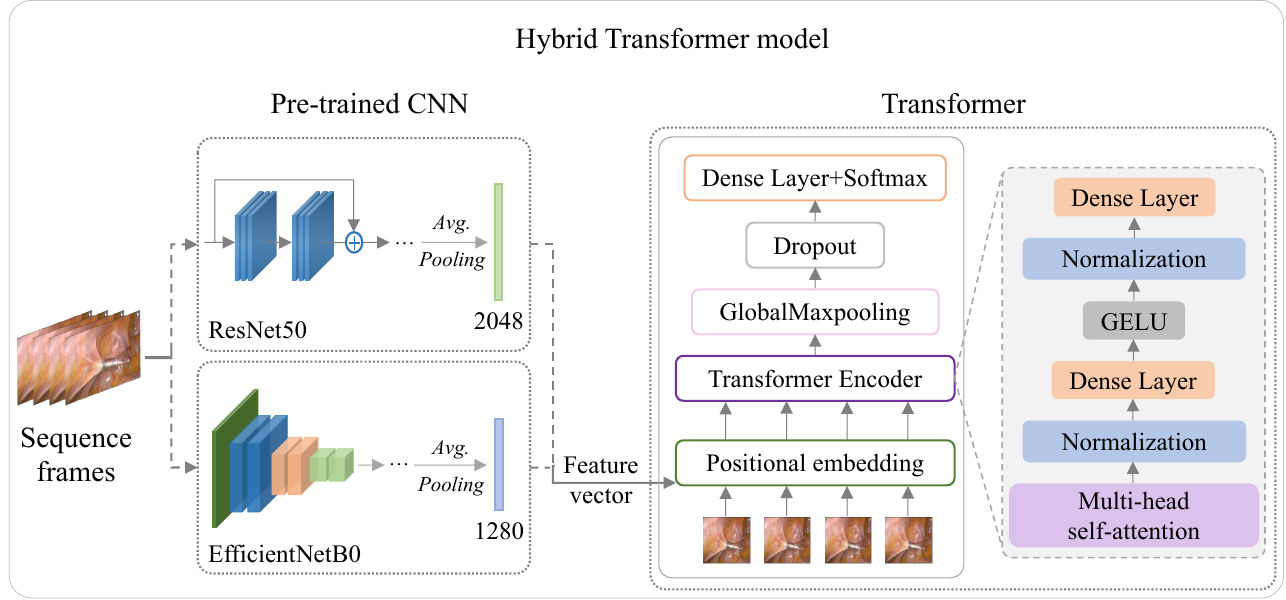}
    \caption{Proposed hybrid transformer architecture.} 
    \label{figure:2}
\end{figure*}

The fusion of convolutional neural networks (CNNs) with transformers has emerged as a compelling approach for advancing surgical event recognition. This hybrid architecture capitalizes on the strengths of both components to achieve robust and accurate recognition of complex surgical events from video data. CNNs excel in extracting spatial features from individual frames, enabling them to capture essential visual information within each frame. These features contribute to the network's understanding of local patterns, textures, instruments, and anatomical structures present in laparoscopic videos. 
Integrating transformers into this architecture addresses the temporal dimension. Transformers' self-attention mechanisms enable them to learn global dependencies and long-range temporal associations within video sequences. By aggregating information from various frames, transformers capture the temporal context critical for recognizing the flow and progression of surgical events. Accordingly, we propose a hybrid transformer model for event recognition depicted in Fig. ~\ref{figure:2}.

In our model, the CNN component functions as a proficient feature extractor, efficiently capturing intricate spatial details within each video frame. These spatial features provide a high-level representation of the input frames' spatial characteristics.
Before passing the feature vectors through the transformer layers, we apply a positional embedding layer to the input frames. This layer adds positional information to each frame in the sequence to help the transformer model understand the temporal order of frames. The feature vectors with positional embeddings are then passed through the transformer encoder layers. These layers are responsible for capturing temporal relationships and dependencies between frames in the video sequence. They perform self-attention over the frames and apply feed-forward layers to process the sequence. After processing the video sequence through the transformer encoder layers, we apply global max-pooling to obtain a fixed-size representation of the entire video sequence. This operation aggregates information from all frames and summarizes it into a single vector. A dropout layer is added for regularization purposes to prevent overfitting to the irrelevant features in the input sequence. Finally, a dense layer followed by a Softmax activation function is used for event classification. This layer takes the fixed-size representation from the global max-pooling layer and produces class probabilities for video classification.

\begin{figure*}[t!]
    \centering
    \includegraphics[width=12cm]{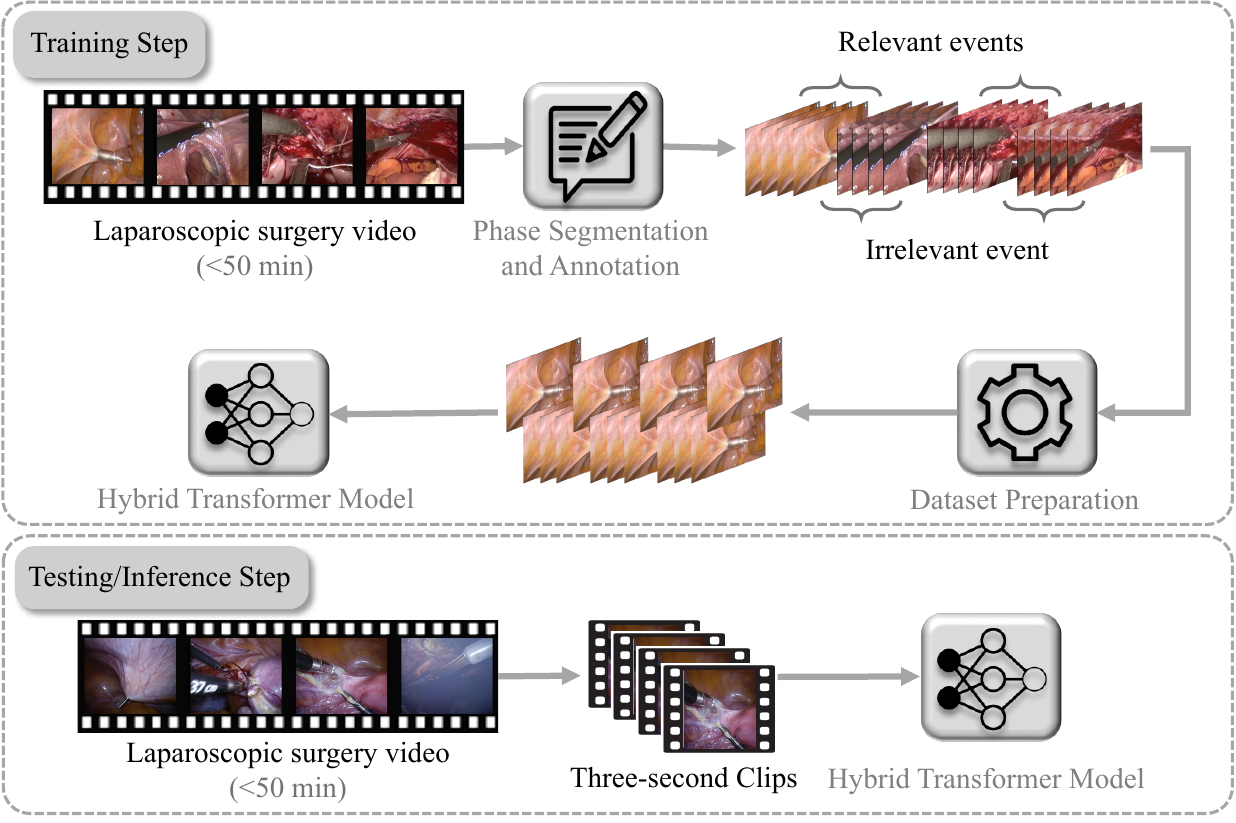}
    \caption{Proposed training and inference framework for event recognition in gynecology laparoscopic videos.} 
    \label{figure:3}
\end{figure*}

\subsection{Training and Inference Framework}
Our dataset comprises laparoscopic surgery videos ranging in duration from 10 minutes to 3 hours, all encoded with a temporal resolution of 30 frames per second (see section \ref{sec:setups}). The surgical events within these videos are temporally segmented and classified into four relevant events and the remaining as irrelevant events. These events span from less than a second to over a minute, depending on the specific surgical procedure and the patient's condition. 
Instead of directly feeding consecutive frames to the network, we adopt a dedicated frame sampling strategy termed \textit{input dropout} to enhance the robustness of trained networks to the surgical content and speed variations. 
Concretely, we split all segments in each class into two-to-three-second video clips with a one-second overlap. We set the input sequence length for the network as ten frames. These frames are randomly selected from each 60-to-90-frame clip. This strategy maximizes variations in the combination of input frames presented to the network while preventing deviation of network parameters by unintended learning of the surgeon's speed. 
We also employ offline data augmentation techniques to negate the problem of class imbalance and increase the diversity of training data. Figure\ref{figure:3} provides a comprehensive depiction of our training and inference framework.

\section{Experimental Setups}
\label{sec:setups}

\paragraph{Data augmentation:}
As mentioned in Section \ref{sec:approach}, we employ various data augmentation techniques to amplify the variety of training data accessible to the model, thereby mitigating overfitting risks and improving model generalization~\cite{ghamsarian2021relevance}. For each video segment in the dataset, We randomly select a set of spatial and non-spatial transformations, apply the same transformation to all frames within the segment, and re-encode the video segment. As spatial augmentations, we apply horizontal flipping with a probability of $0.5$ and Gaussian blur with sigma = 5. Besides, the non-spatial transformations include gamma contrast adjustment with the gamma value being set to $0.5$, brightness changes in the range of $[-0.3, 0.3]$, and saturation changes with a value of $1.5$.

\paragraph{Alternative Methods:} To validate the efficiency of our proposed hybrid transformer model, we proceed to train multiple instances of the CNN-RNN architecture. Concretely, we utilize the ResNet50 and EfficientNetB0 architectures as the backbone feature extraction component, complemented by four distinct variants of recurrent layers: long short-term memory (LSTM), gated recurrent units (GRU), Bidirectional LSTM (BiLSTM), and Bidirectional GRU (BiGRU).

\paragraph{Neural Network Settings: }  In all networks, the backbone architecture employed for feature extraction has been initialized with ImageNet pre-trained parameters. 
Regarding our hybrid transformer model, we make strategic choices for the following key parameters:
\begin{enumerate}
    \item \textit{Embedding Dimension:} We set the embedding dimension to match the number of features extracted from our CNN model. This dimension equals 2048 for "ResNet" and 1280 for "EfficientNetB0". This parameter defines the size of the input vectors that the model processes.
    \item \textit{Multi-Head Attention:} We opt for a configuration with 16 attention heads. Each attention head learns distinct relationships and patterns within the input video data, allowing the model to capture a diverse range of dependencies and variations in the information, enhancing its ability to recognize intricate patterns within laparoscopic videos.
    \item \textit{Classification Head:} The dense layer dimension is set to eight. This choice aims to strike a balance between model capacity and computational efficiency, ensuring that the model can effectively process and transform the information captured by the attention mechanism. At the network's top, we sequentially add a Global-MaxPooling layer, a Dropout layer with a rate of $0.5$, and a Dense layer followed by a Softmax activation.
\end{enumerate}

In our alternative CNN-RNN models, the RNN component is implemented with a single recurrent layer consisting of 64 units. This design enables the model to incorporate high-level representations of features corresponding to individual frames, effectively capturing the subtle temporal dynamics within the video sequence. Following this, we implement dropout regularization with a rate of $ 0.4$ to mitigate overfitting, followed by a Dense Layer composed of 64 units. Subsequently, we incorporate a global average pooling layer, followed by a dropout layer with a rate of $0.5$. Ultimately, we conclude this architectural sequence with a Dense layer, employing a Softmax activation function. 

\paragraph{Training settings:}
The primary objective of this study is to identify and classify four critical events, including abdominal access, bleeding, needle passing, and coagulation/transection, within a diverse collection of gynecologic laparoscopic videos that include a multitude of events. To accomplish this task, we employ a binary classification strategy. In this approach, we divide the input videos into two distinct classes: the specific relevant event we seek to identify and all other events present in the video. This process is iteratively repeated for each of the four relevant events, resulting in the creation of four distinct binary classification models.
In our analysis, the videos are resized to the resolution of $224\times 224$ pixels. We train our hybrid transformer model on our video dataset using binary cross entropy loss function and the Adam optimizer with a learning rate of $\alpha = 0.001$.
We assign $80\%$ of the annotations to the training set and the remaining $20\%$ to the test set.

\paragraph{Evaluation metrics: }
For a comprehensive assessment of action recognition outcomes, we utilize four prominent evaluation metrics that are particularly well-suited for surgical action recognition. These metrics include accuracy, precision, and F1-score.

\section{Experimental Results}
\label{sec:results}

Table~\ref{tbl:label3} presents a comparative analysis of the proposed hybrid transformer's performance against various CNN-RNN architectures based on accuracy and F1-score. Notably, these results are derived from experiments conducted on both the original and the augmented datasets. Overall, the hybrid Transformer models exhibit higher average accuracy when applied to the augmented dataset compared to the original dataset. The experimental results reveal the superiority of the transformer model against the alternative CNN-RNN models when coupled with both ResNet50 and EfficientnetB0 networks. In particular, the combination of ResNet50 and the transformer model achieves the highest average accuracy ($86.10\%$) and F1-score ($86.03\%$), demonstrating its superior capability in identifying events within laparoscopic surgery videos. The second-best performing model comprises the EfficientNetB0 network with the Transformer, achieving an average accuracy of $85.62\%$ and an F1-score of $85.56\%$. 
It's important to highlight that the ResNet50-Transformer model achieves an impressive accuracy of $93.75\%$ in correctly recognizing abdominal access events.

\begin{table*}[t!]
\caption{Quantitative comparisons between the proposed architecture and alternative methods based on accuracy and F1-score. The best results for each case are bolded.}
\label{tbl:label3}
\centering
\resizebox{1\textwidth}{!}{%
\begin{tabular}{l*{7}{>{\centering\arraybackslash}m{2.1cm}}}
\specialrule{.12em}{.05em}{.05em}
\rowcolor{shadecolor}\multicolumn{7}{c}{Original Dataset}\\\specialrule{.12em}{.05em}{.05em}
 & & Abd. Access & Needle Passing& Bleeding & Coag./Tran. & Average\\ \cmidrule(lr){3-3}\cmidrule(lr){4-4}\cmidrule(lr){5-5}\cmidrule(lr){6-6}\cmidrule(lr){7-7}
 Backbone  & Head & Acc | F1 (\%)& Acc | F1 (\%)& Acc | F1 (\%)& Acc | F1 (\%)& Acc | F1 (\%)\\
\specialrule{.12em}{.05em}{.05em}
\multirow{5}{*}{ResNet50} & LSTM & 90.00 | 89.90 & 83.08 | 83.07 & \textbf{77.43} | \textbf{77.40} & \textbf{88.03} | \textbf{87.94} & 84.63 | 84.58 \\
 &GRU & 91.25 | 91.18 & 85.54 | 85.48 & 74.78 | 74.72 & 86.32 | 86.11 & 84.47 | 84.37 \\
 &BiLSTM& 88.75 | 88.61 & 86.15 | 86.15 & 70.35 | 69.86 & 86.75 | 86.66 & 83.00 | 82.82 \\
 &BiGRU & 87.50 | 87.30 & 84.31 | 84.30 & 73.45 | 73.15 & 87.39 | 87.33 & 83.16 | 83.02 \\
 &Transformer & \textbf{92.50} | \textbf{92.46} & \textbf{87.31} | \textbf{87.31} & 74.34 | 73.67 & 86.97 | 86.85 & \textbf{85.25} | \textbf{85.07} \\
 \specialrule{.12em}{.05em}{0.05em}
 \multirow{5}{*}{EfficientNetB0} & LSTM & 86.25 | 86.20 & 85.85 | 85.68 & 75.22 | 74.94 & 86.11 | 85.92 & 83.35 | 83.18 \\
 &GRU & 87.50 | 87.37 & 87.92 | 87.71 & 76.11 | 75.41 & 86.32 | 86.15 & 84.46 | 84.18 \\
 &BiLSTM& \textbf{88.75} | \textbf{88.71} & 87.46 | 87.36 & 74.34 | 73.99 & 86.54 | 86.35 & 84.27 | 84.10 \\
 &BiGRU & 87.50 | 87.43 & 83.62 | 83.23 & \textbf{77.88} | \textbf{77.42} & \textbf{88.89} | \textbf{89.60} & 84.47 | 84.42 \\
 &Transformer & 86.25 | 86.25 & \textbf{88.62} | \textbf{88.54} & 75.66 | 75.58 & 88.46 | 88.37 & \textbf{84.74} | \textbf{84.68}\\
\specialrule{.12em}{.05em}{0.05em}
\rowcolor{shadecolor}\multicolumn{7}{c}{Augmented Dataset}\\\specialrule{.12em}{.05em}{.05em}
& & Abd. Access & Needle Passing& Bleeding & Coag./Tran. & Average\\ \cmidrule(lr){3-3}\cmidrule(lr){4-4}\cmidrule(lr){5-5}\cmidrule(lr){6-6}\cmidrule(lr){7-7}
Backbone  & Head & Acc | F1 (\%)& Acc | F1 (\%)& Acc | F1 (\%)& Acc | F1 (\%)& Acc | F1 (\%)\\
\specialrule{.12em}{.05em}{.05em}
\multirow{5}{*}{ResNet50} & LSTM & 91.20 | 91.18 & 76.00 | 74.88 & 77.60 | 77.27 & 71.43 | 74.88 & 79.06 | 78.38 \\
 &GRU & 90.00 | 89.90 & 75.69 | 75.05 & 74.80 | 74.80 & 67.31 | 63.92 & 76.95 | 75.92 \\
 &BiLSTM& 92.47 | 92.48 & 80.15 | 80.00 & 78.40 | 78.40 & 70.51 | 68.51 & 80.38 | 79.85 \\
 &BiGRU & 91.25 | 91.18 & 77.69 | 77.02 & 78.00 | 77.84 & 71.67 | 70.10 & 79.65 | 79.03 \\
 &Transformer & \textbf{93.75} | \textbf{93.73} & \textbf{80.69} | \textbf{80.67} & \textbf{86.00} | \textbf{85.98} & \textbf{83.97} | \textbf{83.75} & \textbf{86.10} | \textbf{86.03} \\
 \specialrule{.12em}{.05em}{0.05em}
 \multirow{5}{*}{EfficientNetB0} & LSTM & \textbf{92.50} | \textbf{92.48} & \textbf{84.92} | \textbf{84.86} & 80.40 | 80.26 & 71.15 | 69.80 & 82.24 | 81.85 \\
 &GRU & 88.75 | 88.71 & 80.46 | 80.17 & 82.00 | 81.82 & 66.03 | 62.08 & 79.31 | 78.19 \\
 &BiLSTM& 90.00 | 89.94 & 82.38 | 82.15 & 81.60 | 81.60 & 72.01 | 70.07 & 81.50 | 80.94 \\
 &BiGRU & 91.25 | 91.22 & 83.31 | 83.02 & 83.60 | 83.46 & 72.65 | 71.45 & 82.70 | 82.32 \\
 &Transformer & 88.75 | 88.73 & 82.85 | 82.83 & \textbf{85.20} | \textbf{85.11} & \textbf{85.68} | \textbf{85.56} & \textbf{85.62} | \textbf{85.56} \\
\specialrule{.12em}{.05em}{0.05em}

\end{tabular}
}
\end{table*}

\begin{figure*}[t!]
    \centering
    \includegraphics[width=0.8\textwidth]{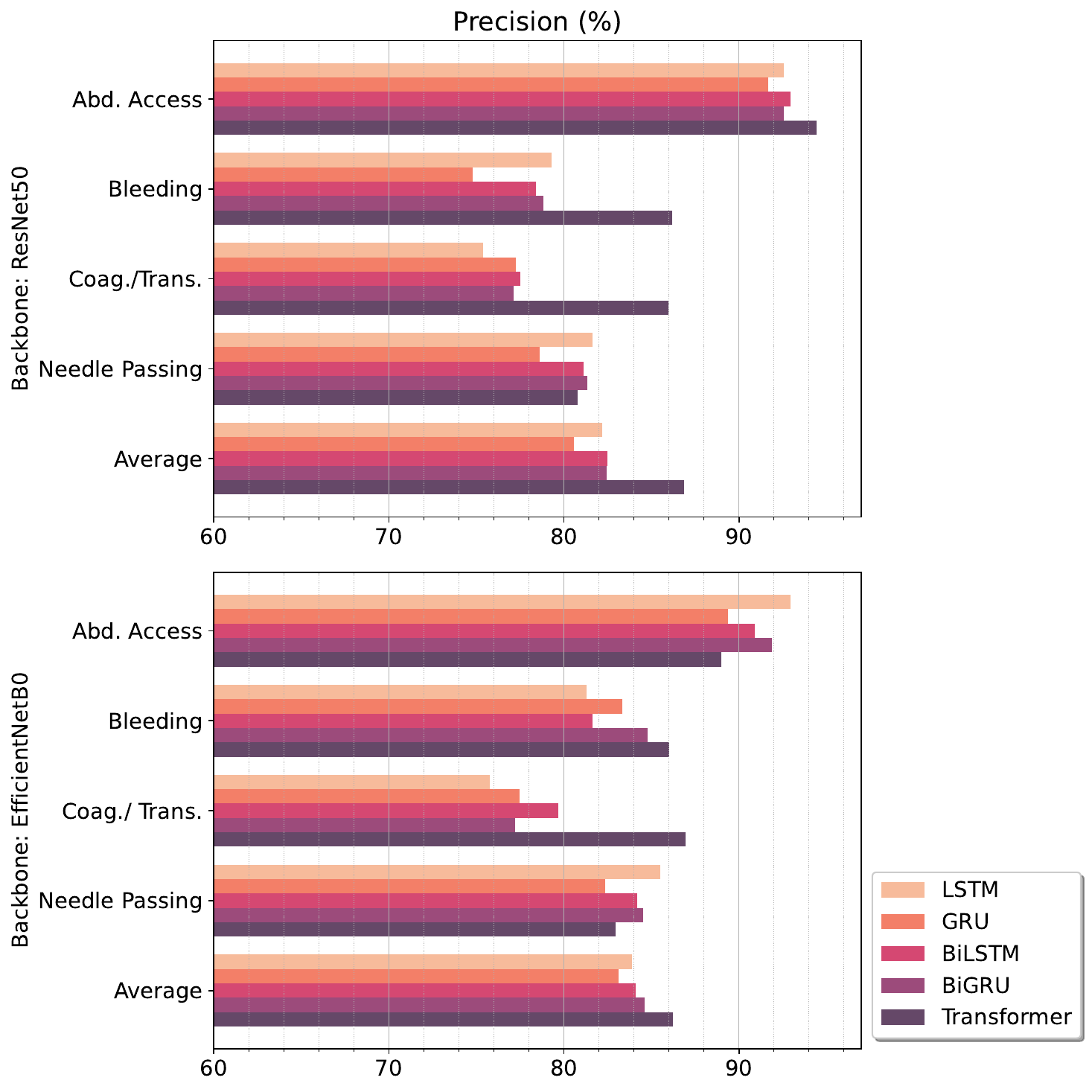}
    \caption{Comparisons between the precision achieved by the proposed method versus alternative methods for relevant event recognition.} 
    \label{figure:4}
\end{figure*}

For further evaluation, we compared event precision across various network architectures, as illustrated in Figure~\ref{figure:4}. In the case of ResNet50, it's evident that the abdominal access event stands out with remarkable precision when paired with the transformer model, achieving an impressive $94.44\%$ accuracy. When considering the overall average precision, the ResNet50-Transformer combination emerges as the top performer. With the EfficientNetB0 backbone, the LSTM network demonstrates the highest precision for the abdominal access event, reaching an accuracy rate of $92.93\%$. Nevertheless, when examining the average precision across all events, the Transformer model consistently outperforms the other networks, showcasing its superiority in event recognition tasks.

\begin{figure*}[t!]
    \centering
    \includegraphics[width=0.95\textwidth]{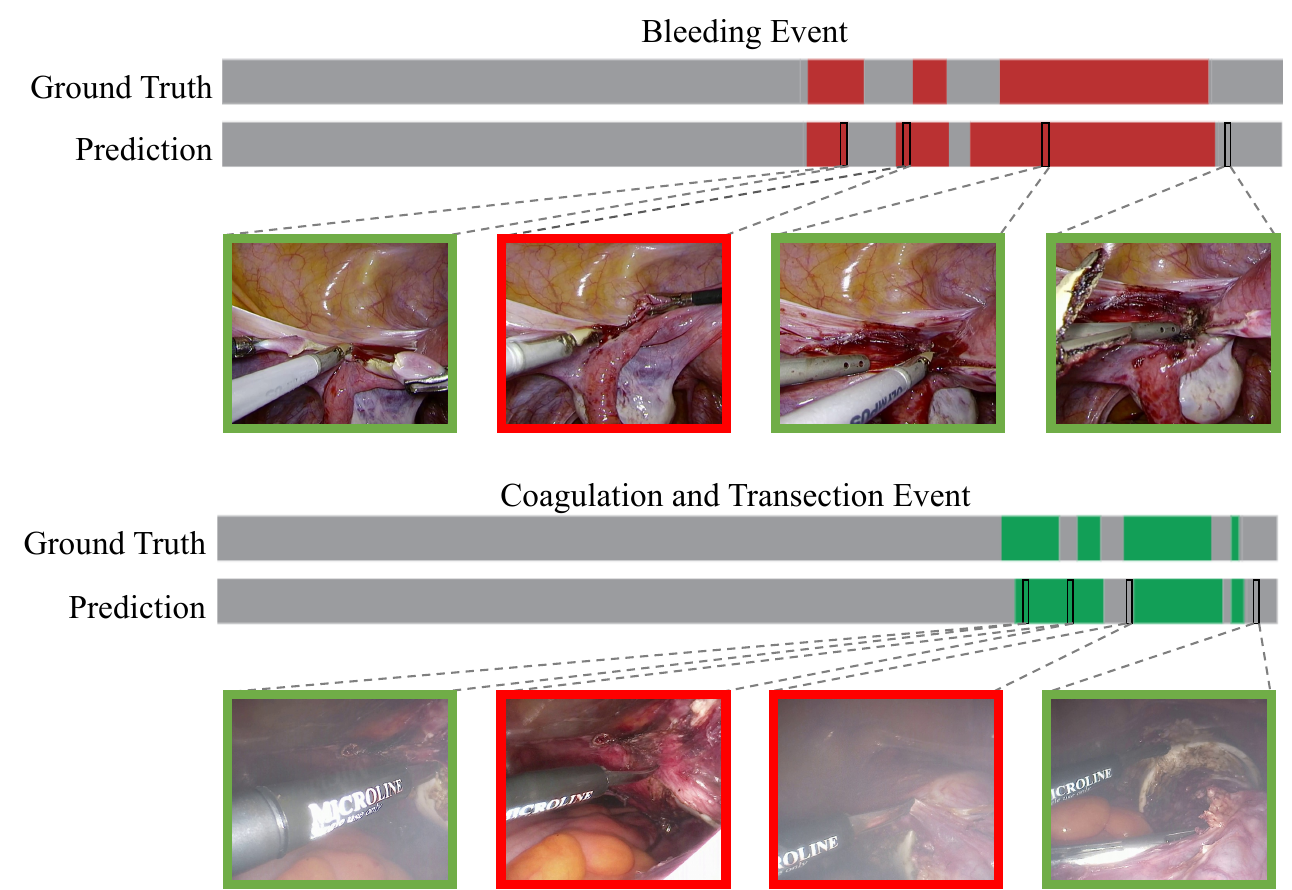}
    \caption{Visualization of the proposed method's (ResNet50-Transformer) performance for a representative challenging segment of a laparoscopic video. The correct and incorrect predictions are shown with green and red borders, respectively.} 
    \label{figure:5}
\end{figure*}

Figure~\ref{figure:5} illustrates segments featuring two challenging events, Bleeding and Coagulation/Transection, recognized by our proposed ResNet50-Transformer model. The first segment consists of roughly 8 minutes from a laparoscopic surgery video, including various instances of bleeding events.  
Four randomly selected representative frames from relevant or irrelevant events are depicted to showcase the challenging task of relevant event recognition in these videos. Remarkably, our model successfully identifies almost all bleeding events within the video. This achievement is significant given the inherent challenge of detecting bleeding, as it can occur during various surgical phases. The model's capability to recognize flowing blood, irrespective of the surgical instruments in use, underscores its robust performance. Of course, distinguishing between flowing and non-flowing blood can be quite complicated and lead to "false positive" detections, as demonstrated for the second visualized frame of the first sequence. 
Regarding the coagulation/transection events, it should be noted that laparoscopic videos can potentially contain smoky scenes, which increases the complexity of detecting this particular event. Despite these challenges, our model performs successfully in recognizing Coagulation/Transection events. However, in some instances, specific video clips annotated as irrelevant events are identified by the model as coagulation/transection event. This discrepancy can be attributed to the resemblance between these video clips and the actual coagulation/transection events.

\section{Conclusion}
\label{sec:conclusion}

Identifying relevant events in laparoscopic gynecology videos is a critical step in a majority of workflow analysis applications for this surgery. This paper has presented and evaluated a comprehensive dataset for recognizing four relevant events linked to significant challenges and complications in laparoscopic gynecology surgery.
Furthermore, we have proposed a training/inference framework featuring a hybrid-transformer-based architecture tailored for the challenging task of event recognition within this surgery. Through extensive experiments employing different pre-trained CNN networks, in combination with the transformers or recurrent layers, we have identified the optimal architecture for our classification objective.

\subsubsection*{Acknowledgements} 
This work was funded by the FWF Austrian Science Fund under grant
P 32010-N38. The authors would like to thank Daniela Stefanics for her help with data annotations.

\bibliographystyle{splncs04}
\bibliography{references.bib}

\begin{thebibliography}{10}
\providecommand{\url}[1]{\texttt{#1}}
\providecommand{\urlprefix}{URL }
\providecommand{\doi}[1]{https://doi.org/#1}

\bibitem{aldahoul2021transfer}
Aldahoul, N., Karim, H.A., Tan, M.J.T., Fermin, J.L.: Transfer learning and
  decision fusion for real time distortion classification in laparoscopic
  videos. IEEE Access  \textbf{9},  115006--115018 (2021)

\bibitem{bello2019attention}
Bello, I., Zoph, B., Vaswani, A., Shlens, J., Le, Q.V.: Attention augmented
  convolutional networks. In: Proceedings of the IEEE/CVF international
  conference on computer vision. pp. 3286--3295 (2019)

\bibitem{czempiel2020tecno}
Czempiel, T., Paschali, M., Keicher, M., Simson, W., Feussner, H., Kim, S.T.,
  Navab, N.: Tecno: Surgical phase recognition with multi-stage temporal
  convolutional networks. In: Medical Image Computing and Computer Assisted
  Intervention--MICCAI 2020: 23rd International Conference, Lima, Peru, October
  4--8, 2020, Proceedings, Part III 23. pp. 343--352. Springer (2020)

\bibitem{dosovitskiy2020image}
Dosovitskiy, A., Beyer, L., Kolesnikov, A., Weissenborn, D., Zhai, X.,
  Unterthiner, T., Dehghani, M., Minderer, M., Heigold, G., Gelly, S., et~al.:
  An image is worth 16x16 words: Transformers for image recognition at scale.
  arXiv preprint arXiv:2010.11929  (2020)

\bibitem{funke2019video}
Funke, I., Mees, S.T., Weitz, J., Speidel, S.: Video-based surgical skill
  assessment using 3d convolutional neural networks. International journal of
  computer assisted radiology and surgery  \textbf{14},  1217--1225 (2019)

\bibitem{ghamsarian2020enabling}
Ghamsarian, N.: Enabling relevance-based exploration of cataract videos. In:
  Proceedings of the 2020 International Conference on Multimedia Retrieval. pp.
  378--382 (2020)

\bibitem{ghamsarian2020relevance}
Ghamsarian, N., Amirpourazarian, H., Timmerer, C., Taschwer, M.,
  Sch{\"o}ffmann, K.: Relevance-based compression of cataract surgery videos
  using convolutional neural networks. In: Proceedings of the 28th ACM
  International Conference on Multimedia. pp. 3577--3585 (2020)

\bibitem{ghamsarian2021lensid}
Ghamsarian, N., Taschwer, M., Putzgruber-Adamitsch, D., Sarny, S., El-Shabrawi,
  Y., Schoeffmann, K.: Lensid: a cnn-rnn-based framework towards lens
  irregularity detection in cataract surgery videos. In: Medical Image
  Computing and Computer Assisted Intervention--MICCAI 2021: 24th International
  Conference, Strasbourg, France, September 27--October 1, 2021, Proceedings,
  Part VIII 24. pp. 76--86. Springer (2021)

\bibitem{ghamsarian2021relevance}
Ghamsarian, N., Taschwer, M., Putzgruber-Adamitsch, D., Sarny, S., Schoeffmann,
  K.: Relevance detection in cataract surgery videos by spatio-temporal action
  localization. In: 2020 25th International Conference on Pattern Recognition
  (ICPR). pp. 10720--10727. IEEE (2021)

\bibitem{golany2022artificial}
Golany, T., Aides, A., Freedman, D., Rabani, N., Liu, Y., Rivlin, E., Corrado,
  G.S., Matias, Y., Khoury, W., Kashtan, H., et~al.: Artificial intelligence
  for phase recognition in complex laparoscopic cholecystectomy. Surgical
  Endoscopy  \textbf{36}(12),  9215--9223 (2022)

\bibitem{he2022empirical}
He, Z., Mottaghi, A., Sharghi, A., Jamal, M.A., Mohareri, O.: An empirical
  study on activity recognition in long surgical videos. In: Machine Learning
  for Health. pp. 356--372. PMLR (2022)

\bibitem{huang2022surgical}
Huang, G.: Surgical action recognition and prediction with transformers. In:
  2022 IEEE 2nd International Conference on Software Engineering and Artificial
  Intelligence (SEAI). pp. 36--40. IEEE (2022)

\bibitem{jin2020multi}
Jin, Y., Li, H., Dou, Q., Chen, H., Qin, J., Fu, C.W., Heng, P.A.: Multi-task
  recurrent convolutional network with correlation loss for surgical video
  analysis. Medical image analysis  \textbf{59},  101572 (2020)

\bibitem{kiyasseh2023vision}
Kiyasseh, D., Ma, R., Haque, T.F., Miles, B.J., Wagner, C., Donoho, D.A.,
  Anandkumar, A., Hung, A.J.: A vision transformer for decoding surgeon
  activity from surgical videos. Nature Biomedical Engineering pp. 1--17 (2023)

\bibitem{leibetseder2017image}
Leibetseder, A., Primus, M.J., Petscharnig, S., Schoeffmann, K.: Image-based
  smoke detection in laparoscopic videos. In: Computer Assisted and Robotic
  Endoscopy and Clinical Image-Based Procedures: 4th International Workshop,
  CARE 2017, and 6th International Workshop, CLIP 2017, Held in Conjunction
  with MICCAI 2017, Qu{\'e}bec City, QC, Canada, September 14, 2017,
  Proceedings 4. pp. 70--87. Springer (2017)

\bibitem{lim2017laparoscopic}
Lim, S., Ghosh, S., Niklewski, P., Roy, S.: Laparoscopic suturing as a barrier
  to broader adoption of laparoscopic surgery. JSLS: Journal of the Society of
  Laparoendoscopic Surgeons  \textbf{21}(3) (2017)

\bibitem{loukas2018video}
Loukas, C.: Video content analysis of surgical procedures. Surgical endoscopy
  \textbf{32},  553--568 (2018)

\bibitem{loukas2015smoke}
Loukas, C., Georgiou, E.: Smoke detection in endoscopic surgery videos: a first
  step towards retrieval of semantic events. The International Journal of
  Medical Robotics and Computer Assisted Surgery  \textbf{11}(1),  80--94
  (2015)

\bibitem{loukas2018keyframe}
Loukas, C., Varytimidis, C., Rapantzikos, K., Kanakis, M.A.: Keyframe
  extraction from laparoscopic videos based on visual saliency detection.
  Computer methods and programs in biomedicine  \textbf{165},  13--23 (2018)

\bibitem{lux2010novel}
Lux, M., Marques, O., Sch{\"o}ffmann, K., B{\"o}sz{\"o}rmenyi, L., Lajtai, G.:
  A novel tool for summarization of arthroscopic videos. Multimedia Tools and
  Applications  \textbf{46},  521--544 (2010)

\bibitem{namazi2018automatic}
Namazi, B., Sankaranarayanan, G., Devarajan, V.: Automatic detection of
  surgical phases in laparoscopic videos. In: Proceedings on the international
  conference in artificial intelligence (ICAI). pp. 124--130 (2018)

\bibitem{namazi2022contextual}
Namazi, B., Sankaranarayanan, G., Devarajan, V.: A contextual detector of
  surgical tools in laparoscopic videos using deep learning. Surgical endoscopy
  pp. 1--10 (2022)

\bibitem{Nasirihaghighi2023}
Nasirihaghighi, S., Ghamsarian, N., Stefanics, D., Schoeffmann, K., Husslein,
  H.: Action recognition in video recordings from gynecologic laparoscopy. In:
  2023 IEEE 36th International Symposium on Computer-Based Medical Systems
  (CBMS). pp. 29--34 (2023)

\bibitem{AAiLS}
Polat, M., Incebiyik, A., Tammo, O.: Abdominal access in laparoscopic surgery
  of obese patients: a novel abdominal access technique. Annals of Saudi
  Medicine  \textbf{43}(4),  236--242 (2023)

\bibitem{schoeffmann2015keyframe}
Schoeffmann, K., Del~Fabro, M., Szkaliczki, T., B{\"o}sz{\"o}rmenyi, L.,
  Keckstein, J.: Keyframe extraction in endoscopic video. Multimedia Tools and
  Applications  \textbf{74},  11187--11206 (2015)

\bibitem{shi2022recognition}
Shi, C., Zheng, Y., Fey, A.M.: Recognition and prediction of surgical gestures
  and trajectories using transformer models in robot-assisted surgery. In: 2022
  IEEE/RSJ International Conference on Intelligent Robots and Systems (IROS).
  pp. 8017--8024. IEEE (2022)

\bibitem{shi2022attention}
Shi, P., Zhao, Z., Liu, K., Li, F.: Attention-based spatial--temporal neural
  network for accurate phase recognition in minimally invasive surgery:
  feasibility and efficiency verification. Journal of Computational Design and
  Engineering  \textbf{9}(2),  406--416 (2022)

\bibitem{twinanda2016endonet}
Twinanda, A.P., Shehata, S., Mutter, D., Marescaux, J., De~Mathelin, M., Padoy,
  N.: Endonet: a deep architecture for recognition tasks on laparoscopic
  videos. IEEE transactions on medical imaging  \textbf{36}(1),  86--97 (2016)

\bibitem{vaswani2017attention}
Vaswani, A., Shazeer, N., Parmar, N., Uszkoreit, J., Jones, L., Gomez, A.N.,
  Kaiser, {\L}., Polosukhin, I.: Attention is all you need. Advances in neural
  information processing systems  \textbf{30} (2017)

\bibitem{wang2018smoke}
Wang, C., Cheikh, F.A., Kaaniche, M., Elle, O.J.: A smoke removal method for
  laparoscopic images. arXiv preprint arXiv:1803.08410  (2018)

\end{thebibliography}

\end{document}